\def\year{2020}\relax
\documentclass[letterpaper]{article} % DO NOT CHANGE THIS
\usepackage{aaai20}  % DO NOT CHANGE THIS
\usepackage{times}  % DO NOT CHANGE THIS
\usepackage{helvet} % DO NOT CHANGE THIS
\usepackage{courier}  % DO NOT CHANGE THIS
\usepackage[hyphens]{url}  % DO NOT CHANGE THIS
\usepackage{graphicx} % DO NOT CHANGE THIS
\usepackage{graphicx}  % DO NOT CHANGE THIS
\usepackage{url}            % simple URL typesetting
\usepackage{booktabs}       % professional-quality tables
\usepackage{amsfonts}       % blackboard math symbols
\usepackage{nicefrac}       % compact symbols for 1/2, etc.
\usepackage{microtype}      % microtypography
\usepackage{graphicx}

\usepackage{enumitem}
\usepackage{amsmath}
\usepackage{comment}
\usepackage{paralist}

\usepackage{color}

\title{Multi-level Head-wise Match and Aggregation in Transformer \\
 for Textual Sequence Matching}
\author{Shuohang Wang\textsuperscript{\rm 1}\textsuperscript{\rm 2}\thanks{Work done when the author was at Singapore Management University}, Yunshi Lan\textsuperscript{\rm 1}, Yi Tay\textsuperscript{\rm 3}\thanks{Now working at Google}, Jing Jiang\textsuperscript{\rm 1}, Jingjing Liu\textsuperscript{\rm 2}\\ 
\textsuperscript{\rm 1}Singapore Management University, %If you have multiple authors and multiple affiliations
\textsuperscript{\rm 2}Microsoft Dynamics 365 AI Research,
\textsuperscript{\rm 3}Nanyang Technological University\\
\{shwang.2014, yslan.2015, jingjiang\}@smu.edu.sg, ytay017@e.ntu.edu.sg, jingjl@microsoft.com % email address must be in roman text type, not monospace or sans serif
}

\begin{document}

\maketitle
\begin{abstract}
Transformer has been successfully applied to many natural language processing tasks.
However, for textual sequence matching, simple matching between the representation of a pair of sequences might bring in unnecessary noise. 
In this paper, we propose a new approach to sequence pair matching with Transformer, by learning head-wise matching representations on multiple levels.
Experiments show that our proposed approach can achieve new state-of-the-art performance on multiple tasks that rely only on pre-computed sequence-vector-representation, such as SNLI, MNLI-match, MNLI-mismatch, QQP, and SQuAD-binary.

\end{abstract}

\section{Introduction}

Textual sequence matching is important for many natural language processing tasks, such as textual entailment~\cite{bowman:emnlp15}, paraphrase identification~\cite{dolan2005automatically}, question answering~\cite{tan2015lstm}, etc. 
There has been a large amount of work focusing on this problem, from exploring classic human-crafted features~\cite{wan2006using}, to tree-based neural structures ~\cite{mou2015:emnlp,tai2015improved}, and a large number of attention-based models~\cite{rock:ICLR2016,yin2015abcnn,parikh2016decomposable,lin2017structured}.
Table~\ref{tbl:example} provides two examples of real sequence matching problems on duplicated question detection and text entailment.

Across the rich history of sequence matching research, there have been two main streams of studies for solving this problem\footnote{As classified in the leaderboard of SNLI, https://nlp.stanford.edu/projects/snli/.}. The first utilizes a sequence encoder (e.g., LSTM/CNN) to obtain static low-dimensional vector representations of sequence pairs. Subsequently, a parameterized function (e.g., Multi-layered Perceptron) is applied on top of the representations to learn a matching score~\cite{bowman:emnlp15,mou2015:emnlp,tai2015improved,lin2017structured}. The second direction learns to aggregate word and phrase level interactions using cross-sentence attention, learning the entire matching function in an end-to-end fashion~\cite{rock:ICLR2016,wang:NAACL2016,yin2015abcnn,devlin2018bert}.

\begin{table}[t]
\centering
\begin{small}
\begin{tabular}{lp{6.5cm}}
\toprule
Q1 & Is there a reason why we should travel alone?            \\
Q2 & What are,some reasons to travel alone?                   \\
L  & is\_duplicated                                               \\
\midrule
\midrule
P & Two young children in blue jerseys, one with the number 9 and one with the number 2 are standing on the wooden steps in a bathroom and washing their hands in a sink.      \\
H & Two kids in numbered jerseys want their hands. \\
L & Entailment   \\
\bottomrule
\end{tabular}
\end{small}
\caption{Examples from Quora Question Pairs (QQP) dataset for duplicated question detection and Stanford Natural Language Inference dataset for text entailment. ``Q1" and ``Q2" are the question pair. ``P" represents the premise and ``H" the hypothesis. ``L'' is the label.}
\label{tbl:example}
\end{table}

While models that utilize cross-sentence attention typically achieve better performance~\cite{devlin2018bert}, the encoders are unfortunately not re-usable. 
Hence, new sequence pairs require re-computation of word/phrasal alignments. 
Consequently, this incurs additional computation costs. 
Conversely, models that learn static representations enjoy reuse of their fixed vector representations and are therefore inherently more efficient (i.e., we only need to compute the scoring function on top of the pre-computed representations).
As pointed out by \citeauthor{reimers2019sentence} \shortcite{reimers2019sentence}, finding the most similar sentence pair in a collection of n = 10,000 sentences need to run original BERT $n\times(n-1)/2 = 49,995,000$ times. 
By making use of static representations, we only need to run BERT n times together with some much cheaper matrix multiplications, which will reduce the running time from 65 hours to 5 seconds.
Moreover, this inherent scalability benefit enables querying of up to millions of documents~\cite{das2018multi} or answers~\cite{seo2018phrase}, which provides a greater recall/coverage in open domain question answering. 
Our work, inspired by this advantage, focuses on improving techniques based on learning with pre-computed representations. 

Notably, there are several well-established methods for matching pre-computed vector representations (e.g., cosine similarity~\cite{tan2015lstm}, element-wise subtraction~\cite{tai2015improved} and direct concatenation followed by Multi-layered Perceptron~\cite{bowman:emnlp15}). 
Although these matching methods have demonstrated reasonable success for LSTM~\cite{tai2015improved} or CNN encoders~\cite{mou2015:emnlp}, they are not ideal for matching representations obtained from Transformer encoders~\cite{vaswani2017attention}. 
Different from other types of encoders, the multi-headed self-attention mechanism that lives at the heart of the Transformer model requires special treatment. 
To this end, how to effectively aggregate information from multiple heads and multiple layers of the Transformer remains an unanswered research question. 

This work is mainly concerned with (1) empirically studying the behavior of the multi-head attention in pretrained Transformer models~\cite{devlin2018bert} and (2) proposing a new matching aggregation scheme based on our empirical observations. 
More specifically, based on our empirical analysis and visualization of the behavior of the underlying multi-headed attention, we arrive at several conclusions. 
Firstly, most of the heads will assign higher attention weights to specific words. 
Secondly, heads from different layers target at different aspects of the sequence. 
Thirdly, some heads pay attention to the same or related words. 
Lastly, some heads pay attention to stopwords or punctuation marks. 
Intuitively, it seems as though the representation ability of the Transformer encoder is spread out across the network, heads, and layers. 
Therefore, specialized aggregation is necessary.

As per our observations, different heads may cover different aspects of information. 
Therefore, mixing the heads together with a fully connected layer may be sub-optimal. 
Instead, we propose matching the corresponding heads from different sequences independently and then aggregating them for sequence matching. 
This is in a similar spirit to the compare-aggregate framework with cross-sequence attention~\cite{wang:NAACL2016,parikh2016decomposable}, albeit largely based on the head level representations.

\paragraph{Our contributions} The contributions of our work can be summarized as follows:
\begin{compactitem}
    \item We propose a new head-wise matching method, specifically designed for the Transformer model~\cite{vaswani2017attention,devlin2018bert}.
    \item We investigate the utility of different matching functions for head-wise matching. Based on our experiments, we find that element-wise matching representation works the best for head-wise matching.
    \item We show that we are able to further boost the performance by making use of the head matching in different layers of Transformer.
    \item  We explore different methods to integrate the head-wise matching representation, arriving at the conclusion that max pooling based method works the best.
    \item We achieve state-of-the-art results on multiple tasks, including SNLI~\cite{bowman:emnlp15}, QQP~\cite{wang2018glue}, MNLI~\cite{N18-1101}, SQuAD-binary~\cite{rajpurkar2018know}, relying on sentence vector representations.
\end{compactitem}

\section{Related Work}

Learning the relationship between textual sequence pairs is a fundamental problem in natural language processing~(NLP). A wide range of NLP tasks fit into this problem formulation, ranging from textual entailment~\cite{bowman:emnlp15} to question answering~\cite{tan2015lstm}. 

The dominant approaches to sequence matching are largely based on cross-sentence attention~\cite{wang:NAACL2016,parikh2016decomposable,rock:ICLR2016}. The key idea is to learn word/phrasal alignment between sequence pairs. Models that compute such alignment on the fly can perform well on many benchmarks. However, the inability to reuse representations hampers its usability in real-world applications.  

To mitigate this issue, pre-computed vector representation can be used, due to its ability for reuse. \citeauthor{seo2018phrase} \shortcite{seo2018phrase} proposed phrase-indexed question answering~(PIQA), a special setting for reading comprehension that prohibits cross-sentence attention being computed on the fly. Instead, representations should be pre-computed and ``cached''. This provides an intrinsic scalability benefit, enabling efficiency during inference. In practice, requiring the re-computation of alignment on the fly is not suitable for many real-world applications.  \citeauthor{das2018multi} \shortcite{das2018multi} proposed multi-step reasoning using pre-computed vector representations, which can accelerate searching through documents.

Fine-tuning is another option~\cite{devlin2018bert}. The key idea is to pretrain a Transformer model on a large Web corpus with unsupervised objectives. These large pre-trained models are then fine-tuned for downstream tasks. The state-of-the-art pre-trained models such as BERT~\cite{devlin2018bert} and GPT~\cite{radford2018improving} all fine-tune over sequence pairs, and apply self-attention across concatenated pairs. Effectively, these models can compute alignment on the fly, but are expensive for real world applications due to the need of computing token-level cross attention in each forward pass. 
In this paper, we will investigate the usage of these pre-trained models as static encoders.

\section{Model}
\begin{figure*}[t]
\centering
\includegraphics[width=5.5in]{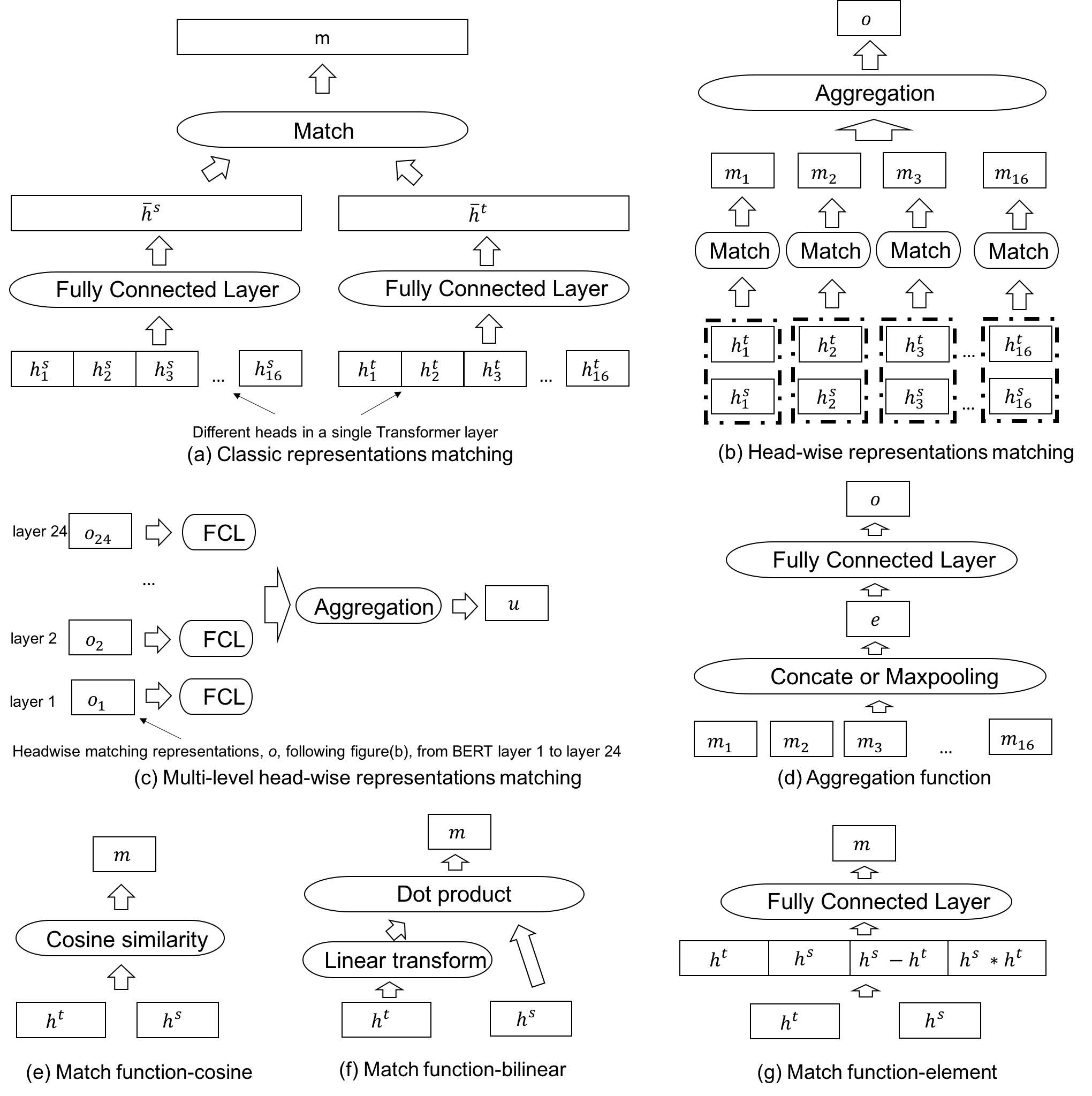}
\caption{An overview of the model. Fully connected layer (FCL) is a non-linear transformation layer. Figures (e)(f)(g) are different Match functions applied in the first three models.}
\label{fig:model}
\end{figure*}

In this section, we will introduce the details of our model. An overview of model architecture is shown in Figure~\ref{fig:model}.
\subsection{Transformer Encoder}
In this paper, we will focus on making use of the vectorized representations built by Transformer~\cite{vaswani2017attention}, which is initialized by BERT~\cite{devlin2018bert}, for sequence matching.

Considering the reusability of encoders, we will not directly concatenate the sequences and run a single Transformer on it for matching.
Instead, without cross-sequence attention, we run Transformer on sequence pairs independently and only match sequences based on their vectorized representations.
This setting is useful for large-scale answer extraction~\cite{seo2018phrase} and multi-step reasoning~\cite{das2018multi}.

Each Transformer layer is constructed by multi-head self-attention~\cite{lin2017structured}:
\begin{equation}
    \text{MultiHead}(\mathbf{H}) = f\left( \text{Concat}\left( \textbf{head}_1, ..., \textbf{head}_{\text{I}}\right) \right),
    \label{eqn:multiheads}
\end{equation}
where $\mathbf{H}$ is the hidden representation from last layer.
$f(.)$ is a non-linear transformation, and the function $\text{Concat}(\cdot,\cdot)$ is to concatenate all the $\text{head}_i$, which is collected by attention mechanism:
\begin{eqnarray}
    \textbf{head}_i = \text{Attention}(\mathbf{H} \mathbf{W}_i^{\text{Q}},\mathbf{H} \mathbf{W}_i^{\text{K}}, \mathbf{H} \mathbf{W}_i^{\text{V}}), \\
    \text{Attention}(\mathbf{Q},\mathbf{K},\mathbf{V})=\text{ softmax } ( \frac{\mathbf{Q}\mathbf{K}^T}{ \sqrt{d} }) \mathbf{V} ,
\end{eqnarray}
where $\mathbf{W}_i^{\text{Q}}, \mathbf{W}_i^{\text{K}}, \mathbf{W}_i^{\text{V}}$ are the weights to learn and $d$ is the hidden size of $\text{head}_i$ for scaling the attention weights. 
In the default setting of Transformer, the hidden representation of the first token, which is an inserted special token ``[CLS]" for classification, represents the whole sequence as follows:
\begin{equation}
     \overline{\mathbf{h}} = \text{MultiHead}(\mathbf{Q}, \mathbf{K}, \mathbf{V} )[0],
     \label{eqn:hbar}
\end{equation}
where $[0]$ is to select the vector in the first position. 
When we need to match a sequence pair, such as in the example in Table~\ref{tbl:example}, the hidden representations of the pair can be obtained as $\overline{\mathbf{h}}^{\text{s}}$ and $\overline{\mathbf{h}}^{\text{t}}$, respectively, which can be used as the input of a match function to build the matching representation.

\subsection{Head-wise Match}
For the most widely used encoders, such as CNN~\cite{kim:emnlp14}, LSTM~\cite{hochreiter1997long}, tree-LSTM/CNN~\cite{tai2015improved}, the representations of different sequences, $\overline{\mathbf{h}}^s$ and $\overline{\mathbf{h}}^t$, will be concatenated and followed by a MLP layer for classification, as shown in Figure~\ref{fig:model}~(a).
While, for the Transformer~\cite{vaswani2017attention}, it is constructed by the self-attention heads.
And each head can focus on a specific aspect of the sequence, as the distribution of self-attention weights are usually quite sharp, as shown in Figure~\ref{fig:visatt}.
This is the special property for Transformer.
In this way, besides directly matching the sequence level representations of Transformer, we can first match the corresponding heads in different sequences and then aggregate the head-wise matching representations to build  the sequence level matching representation.
An overview of the model is shown in Figure~\ref{fig:model}~(b).

Specifically, instead of mixing up different heads to get a better sequence level representation, as shown in Eqn.(\ref{eqn:multiheads}), we directly match the \textbf{head}$_i$ from different sequences.
For simplicity, we use $h_i$ to represent the first vector, which corresponds to the first token of the sequence in the i$^{th}$ head:
\begin{equation}
     \mathbf{h}_i = \textbf{head}_i[0],
\end{equation}
$\mathbf{h}_i$ represents the i aspect of the sequence information. 
Instead of merging the heads from one sequence, we first match all heads, $\mathbf{h}^{\text{s}}_i, \mathbf{h}^{\text{t}}_i$, from the sequence pair:
\begin{equation}
     \mathbf{m}_i = \text{Match}(\mathbf{h}^{\text{s}}_i, \mathbf{h}^{\text{t}}_i)
     \label{eqn:match}
\end{equation}
where $\mathbf{m}_i$ is the head-wise matching representation.
Then another layer is used to aggregate the matched heads:
\begin{equation}
     \mathbf{o} = \text{Aggregation}\left(\mathbf{m}_1, \mathbf{m}_2, ..., \mathbf{m}_{\text{I}}\right),
     \label{eqn:agg1}
\end{equation}
where $\mathbf{o}$ will be used for sequence level matching. The Match and Aggregation functions will be introduced in Eqn.(9-13).

\subsection{Multi-level Head-wise Match}

Different layers of the encoder can represent different levels of information.
Both the shallow and deep structures are useful for the sequence matching.
Moreover, based on our visualization on the attention weights, we find that the heads  in different layers will also pay attention to different aspects of the sequence.
In this way, we can get the head-wise match for every layer in the Transformer.
We use $\mathbf{o}_l$ to represent the head-wise matching representation by making use of all the heads in the k$^{\text{th}}$ layer, as shown in Figure~\ref{fig:model}~(c).
And we aggregate the representations from different layers as follows:
\begin{eqnarray}
    \nonumber
     \mathbf{v}_i &=& \text{ReLU}(\mathbf{o}_i \mathbf{W}^{\text{v}} + \mathbf{b}^{\text{v}}), \\
     \mathbf{u} &=& \text{Aggregation}\left( \mathbf{v}_1, \mathbf{v}_2, ..., \mathbf{v}_k \right),
     \label{eqn:agg2}
\end{eqnarray}
where $\mathbf{W}^v$ and $\mathbf{b}^v$ are the parameters to optimize. 
And the $\mathbf{u}$ is the final representation of the sequence matching, which consists of multi-level head-wise match, which can be used for the final classification.

\subsection{Match and Aggregation Functions}
To the best our knowledge, this is the first work to aggregate the head-wise matching in Transformer for sequence matching.
We have a further explorations on the influence of different head-wise matching functions, in Eqn.~(\ref{eqn:match}), and aggregation functions, in Eqn.~(\ref{eqn:agg1},\ref{eqn:agg2}) , for building the final sequence matching representation.

For the \textbf{match functions} with two vectors, $\mathbf{h}^{\text{s}}, \mathbf{h}^{\text{t}}$, as inputs, the Cosine similarity would the most efficient one without parameters to learn:
\begin{equation}
     m  =  \text{Match}(\mathbf{h}^{\text{s}}, \mathbf{h}^{\text{t}}) = \text{Cosine}(\mathbf{h}^{\text{s}}, \mathbf{h}^{\text{t}}),
     \label{eqn:cosine}
\end{equation}
where the output is a scalar value. 
Another more complicated matching would a bilinear which is also widely used for the representation matching or attention weight computing:
\begin{equation}
     m  =  \text{Match}(\mathbf{h}^{\text{s}}, \mathbf{h}^{\text{t}}) = {(\mathbf{h}^{\text{s}}} \mathbf{W}^{\text{b}})\cdot \mathbf{h}^{\text{t}},
      \label{eqn:bilinear}
\end{equation}
where $\mathbf{W}^{\text{b}}$ is the matrix weights to learn and the output of the match function is a scalar.
The most complicated and widely adopted match function is based on a fully connected layer with element-wise matching~\cite{mou2015:emnlp,tai2015improved,bowman2016fast,bowman2018looking,tay2017compare}:
\begin{eqnarray}
    \nonumber
     \mathbf{m} & = & \text{Match}(\mathbf{h}^{\text{s}}, \mathbf{h}^{\text{t}}), \\
     & = & g\left( \text{Concat}\left( \mathbf{h}^{\text{s}}, \mathbf{h}^{\text{t}},\mathbf{h}^{\text{s}} - \mathbf{h}^{\text{t}},\mathbf{h}^{\text{s}} * \mathbf{h}^{\text{t}} \right) \right),
     \label{eqn:element}
\end{eqnarray}
where the function $g(\cdot)$ is a non-linear transformation with ReLU as activation function and output of this function is a vector.

Next, we also explore two \textbf{aggregation functions}, where the inputs are the different head-wise matching representations, $\mathbf{m}_1,\mathbf{m}_2...\mathbf{m}_{\text{I}}$.
The most efficient way is the max pooling:
\begin{eqnarray}
     \mathbf{e} & = & \text{MaxPooling}(\mathbf{m}_1,\mathbf{m}_2...\mathbf{m}_{\text{I}}),
     \nonumber
     \\
     \mathbf{o} & = & \text{ReLU}(\mathbf{e} \mathbf{W}^{\text{e}} + \mathbf{b}^{\text{e}}),
     \label{eqn:pooling}
\end{eqnarray}
where $\mathbf{W}^{\text{e}}\in \mathbb{R}^{d\times d}$ and $\mathbf{b}^{\text{e}}\in \mathbb{R}^{d}$ are the parameters to learn. 
And the other way is to directly concatenate them and make use of another non-linear transformation layer to map it into a smaller vector representation:
\begin{eqnarray}
     \mathbf{e} & = & \text{Concat}(\mathbf{m}_1,\mathbf{m}_2...\mathbf{m}_{\text{I}}), 
     \nonumber
     \\
     \mathbf{o} & = & \text{ReLU}(\mathbf{e} \mathbf{W}^{\text{c}} + \mathbf{b}^{\text{c}}),
     \label{eqn:concate}
\end{eqnarray}
where $\mathbf{W}^{\text{c}}\in \mathbb{R}^{dI\times d}$ and $\mathbf{b}^{\text{c}}\in \mathbb{R}^{d}$ are the parameters to learn. 
The transformation matrix $\mathbf{W}^{\text{c}}$ here is much larger than $\mathbf{W}^{\text{e}}$.

\subsection{Loss Function}
We will mainly focus on the tasks of sequence pair classification in the paper.
The matching representation $o$ will be used as the input of final loss as follows:
\begin{equation}
     \text{loss}  = -\text{log}\frac{\text{exp}(\mathbf{o}\cdot \mathbf{w}^{\text{label}}+b^{\text{label}})}{\sum_f \text{exp}(\mathbf{o}\cdot \mathbf{w}^{\text{f}}+b^{\text{f}})},
\end{equation}
where $\mathbf{w}^{\text{f}}\in \mathbb{R}^d$ and $\mathbf{b}^{\text{f}}\in \mathbb{R}$ are the parameters to learn.

\section{Experiments}

\begin{table*}[]
\centering
\begin{tabular}{lccccc}
\toprule
 Sentence vector-based models      & SNLI & MNLI-m & MNLI-mm & QQP       & SQuAD-binary \\
\midrule
LSTM~\cite{bowman:emnlp15}   & 80.6 & - &- &- &-\\
SPINN-PI~\cite{bowman2016fast}   & 83.2 &- &- &- &-\\
NSE~\cite{NSE:eacl17}   & 84.6 &- &- &- &-\\
CAFE~\cite{tay2017compare}   & 85.9 &- &- &- &-\\
RSN~\cite{shen2018reinforced}   & 86.3 &- &- &- &-\\
DSA~\cite{yoon2018dynamic}    & 87.4 &-        &-         &-           &-              \\
BiLSTM~\cite{wang2018glue}   &-      & 70.3   & 70.8    & 61.4/81.7 &-              \\
BiLSTM+Cove~\cite{mccann2017learned}   &-      & 64.5   & 64.8    & 59.4/83.3 &-              \\
BiLSTM+ELMo~\cite{peters2018deep}   &-      & 72.9   & 73.4    & 65.6/85.7 &-              \\
Skip-Thought~\cite{kiros2015skip} &-      & 62.9   & 62.8    & 56.4/82.2 &-  \\
InferSent~\cite{conneau2017supervised}  &-      & 58.7  & 59.1    & 59.1/81.7 &-  \\
GenSen~\cite{subramanian2018learning}  &-      & 71.4   & 71.3    & 59.8/82.9 &-  \\
\midrule
Classic Match    & 83.2 & 73.8   & 74.2    & 63.3/85.9 & 61.5         \\
Single-level Head-wise Match   & 87.0 & 77.7   & 77.4    & 67.8/88.1 & 62.1         \\
Multi-level Head-wise Match & \textbf{88.1} & \textbf{79.2}   & \textbf{79.3}    & \textbf{69.0}/\textbf{88.6} & \textbf{62.9}  \\
\midrule \midrule 
SOTA (cross sentence attention)  & 91.1 & 86.7 & 86.0 & 72.4/89.6	 & 83.2 \\

\bottomrule
\end{tabular}
\caption{Experiment results. We only compare with the sentence vector-based models as listed in the SNLI Leaderboard. The results on MNLI-m, MNLI-mm and QQP are tested through GLUE Leaderboard. SOTA are the state-of-the-art models with cross sentence attention on the datasets. }
\label{tbl:expresults}
\end{table*}

\begin{table}
\centering
\begin{tabular}{lccc}
\toprule
             & train   & test    & \#class \\
\midrule
SNLI         & 549,367 & 9,824   & 3       \\
MNLI-m       & 392,702 & 9,796   & 3       \\
MNLI-mm      & 392,702 & 9,847   & 3       \\
QQP          & 363,870 & 390,964 & 2       \\
SQuAD-binary & 130,319 & 11,873  & 2       \\
\bottomrule
\end{tabular}
\caption{The statistics of different datasets. }
\label{tbl:sts}
\end{table}

This section introduces our experiment results, implementation details and further analysis.

\subsection{Datasets}
We test our models on the tasks of 1) Text Entailment, which is to identify the relation (entailment, contradiction and neural) between a sequence pair, such as the datasets of Stanford Natual Language Inference (SNLI)~\cite{bowman:emnlp15}, Multi-Genre Natural Language Inference matched (MNLI-m) and mismatched (MNLI-mm)~\cite{N18-1101}~\footnote{For the setting of MNLI-m, the sequence pairs from training and test sets are derived from the same sources.
For the setting of MNLI-mm, the samples from training and test sets are in different genres. }; 2). Duplicate Question Detection, which is to identify whether the given question pair is duplicate or not, such as Quora Question Pairs (QQP); 3) Question Answering, such as the binary classification setting of Stanford Question Answering Dataset 2.0 (SQuAD-binary)~\cite{rajpurkar2018know} where we only need to predict whether the given passage can answer the question or not.

We follow the setting of GLUE~\footnote{\url{https://gluebenchmark.com/tasks}} to split the datasets of MNLI-m, MNLI-mm and QQP. For SQuAD-binary, we use the pulic dev set as test set.
The statistics of the number of samples and classes of different datasets is shown in Table~\ref{tbl:sts}.

\subsection{Experiment Results}
Our experiment results are shown in Table~\ref{tbl:expresults}.
For the QQP task, we report the performance of F1 and accuracy. 
For the other tasks, we only use the accuracy as the evaluation metric.

We compare our head-wise matching based method with a classic way to matching the representations built by encoders.
The ``Classic Match" is the method making use of the sequence representation built by Transformer, Eqn.(\ref{eqn:hbar}), and the element-wise match function, Eqn.(\ref{eqn:element}), for sequence pair matching.
The ``Single-level Head-wise Match" is our method to aggregate the head-wise matching representation by Eqn.(\ref{eqn:agg1}) with element-wise match function and maxpooling aggregation function.
For a fair comparison, we only use the representations in the final layer for sequence matching for both above-mentioned methods.
Based on our results, we can clearly see that our head-wise matching based method is significantly better than the classic matching model on all the datasets.
We have a further exploration on integrating ``Multi-level Head-wise Match" representations, by Eqn.(\ref{eqn:agg2}), also with element-wise match function and maxpooling aggregation function.
We can also see that it's better than only aggregating the head-wise matching representations in a single layer.

\begin{table*}[t]
\centering
\begin{tabular}{lc}
\toprule
       Match (M) and Aggregate (A) Functions                  & SNLI \\
\midrule
Match\_cosine (Eqn. \ref{eqn:cosine}) + Agg\_concate  (Eqn. \ref{eqn:concate} )         & 86.8 \\
Match\_bilinear (Eqn.  \ref{eqn:bilinear})  +Agg\_concate  (Eqn. \ref{eqn:concate} )           & 86.7 \\
Match\_element (Eqn. \ref{eqn:element}) +Agg\_concate  (Eqn. \ref{eqn:concate} )          & 88.0 \\
Match\_element (Eqn. \ref{eqn:element}) +Agg\_concate (Eqn. \ref{eqn:concate} ) (no hier) & 87.6 \\
Match\_element (Eqn. \ref{eqn:element})+Agg\_maxpooling (Eqn. \ref{eqn:pooling} )           & 88.1 \\
\bottomrule
\end{tabular}
\caption{Comparison of different matching and aggregation functions in the model of Multi-level Head-wise Match. Each matching/aggregate function corresponds to the its actual equation number. ``\textbf{no hier}" is to simply aggregate all the representations for classification, instead of aggregating all the head representations in each layer and then all the layer presentations. }
\label{tbl:expresults2}
\end{table*}

We also compare to other sentence vector-based models: LSTM~\cite{bowman:emnlp15} or BiLSTM~\cite{wang2018glue} are trained from scratch without special initialization;
SPINN-PI~\cite{bowman2016fast} is a stack-augmented parser-interpreter neural network;
NSE~\cite{NSE:eacl17} is based on neural semantic encoder;
CAFE~\cite{tay2017compare} is compare, compress and propagate with alignment-factorized encoder;
RSN~\cite{shen2018reinforced} is a reinforced self-attention encoder;
DSA~\cite{yoon2018dynamic} is a dynamic self-attention encoder.
BiLSTM+Cove~\cite{mccann2017learned} is based a pre-trained encoder on neural machine translation.
BiLSTM+ELMo~\cite{peters2018deep} is based a pre-trained encoder on language modeling.
Skip-Thought~\cite{kiros2015skip}, InferSent~\cite{conneau2017supervised}, GenSen~\cite{subramanian2018learning} are also based on pre-trained encoders trained with different methods.

By comparing all the previous models on matching vectorized sequence representations, we can see that our model achieves the best performance under this setting.
Although our best models are still worse than state-of-the-art models with cross sentence attention, the performance is quite close on the datasets of SNLI and QQP, both of which have much larger training set than MNLI and SQuAD-binary.
For the dataset of SQuAD-binary, all of our models obtain relatively poor performance.
One possible reason is the training set of SQuAD-binary is relatively small and therefore insufficient for learning a good encoder.
Another main reason is that SQuAD is in paragraph level, which is naturally longer than the sequences in SNLI or QQP. As such, the hidden states of Transformer are still not powerful enough to represent all the information of the paragraphs containing more words.

\subsection{Implementation Details}
Our Transformer is initialized by BERT-large~\cite{devlin2018bert}~\footnote{\url{https://github.com/huggingface/pytorch-pretrained-BERT}}, with 24 layers, 16 heads each layer. Each head is a 64 dimentional vector.
We limit the maximal single sequence length of SQuAD-binary to 384 and the other datasets 64.
We set the batch size to be 16.
We tune the learning rate from $[10^{-5},2\times 10^{-5},3\times 10^{-5}]$ and dropout from $[0,0.1,0.2,0.3,0.4]$. 
We use a single GPU, Nvidia V100 with 16G memory, for training the models.

\begin{figure}[t]
\centering
\includegraphics[width=3in]{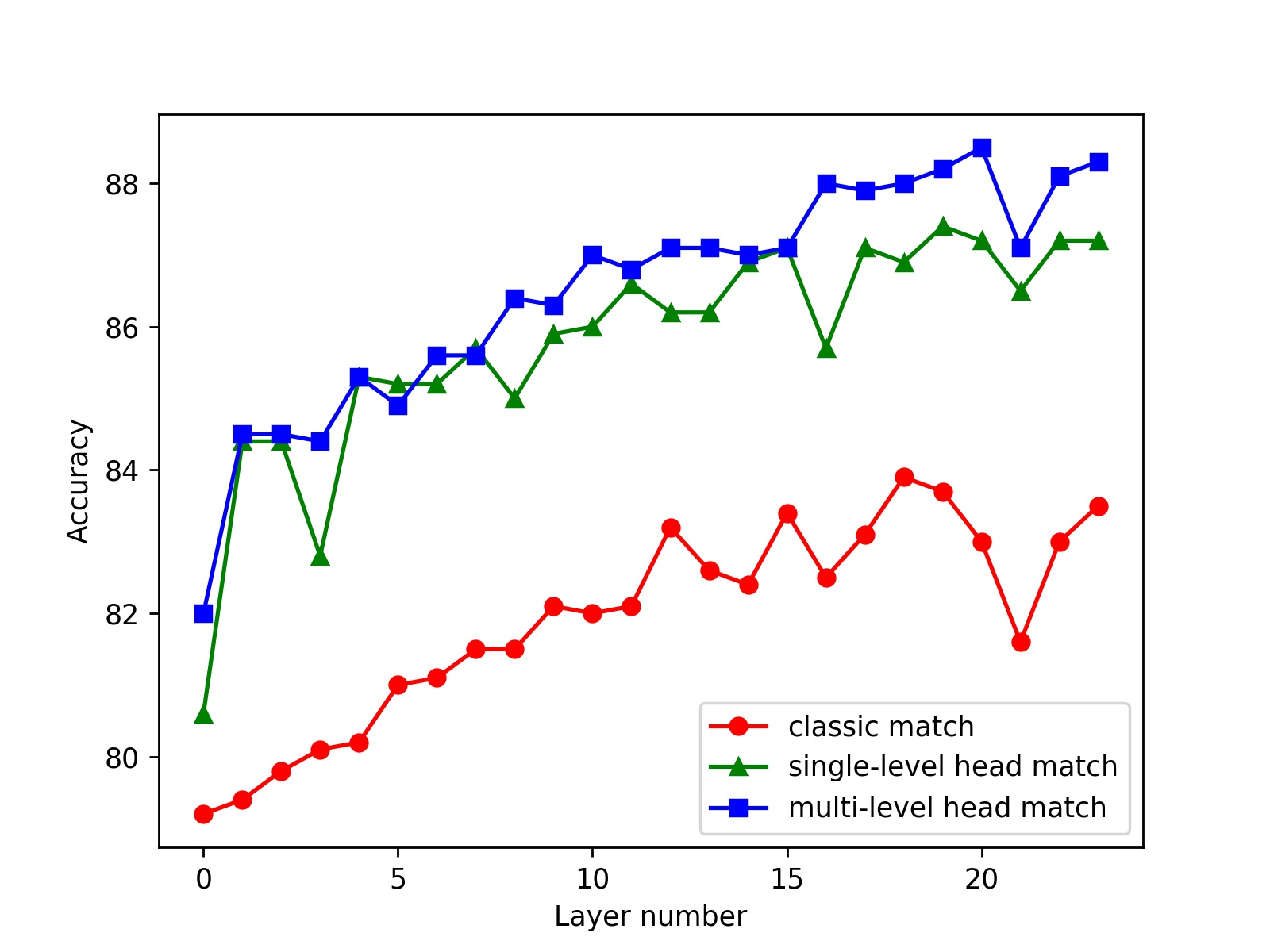}
\caption{The performance of matching models on Transformer with different number of layers. }
\label{fig:headlayers}
\end{figure}
\subsection{Analysis}
This section presents a comprehensive ablative and qualitative analysis.
\paragraph{Effect of Match-Aggregate functions} We analyze the performance of different match functions and aggregation functions based on the experiments on the SNLI dataset. 
The experiment results are shown in Table~\ref{tbl:expresults2}. 
According to our results, we observe  ``Match\_element" performs the best in all of our experiments in Table~\ref{tbl:expresults} also rely on this match function.
Although the ``Match\_cosine" function performs marginally worse, it requires fewer parameters and is also faster than ``Match\_element", serving as a good choice when in large scale settings.

As for the aggregation methods, we can see that ``Agg\_maxpooling" and ``Agg\_concate" achieve similar performance. However, ``Agg\_maxpooling" requires $I$, the number of heads, times fewer parameters for training.
In a similar fashion, we also extend our model with this aggregation method to other datasets.
Additionally, we also have a comparison with the models with and without hierarchical aggregation, denoted as the ``no hier" line in Table~\ref{tbl:expresults2}.
The ``no hier" method performs slightly worse at extracting the most useful head-wise match representation for the final classification.

\begin{figure*}[t]
\centering
\includegraphics[width=6in]{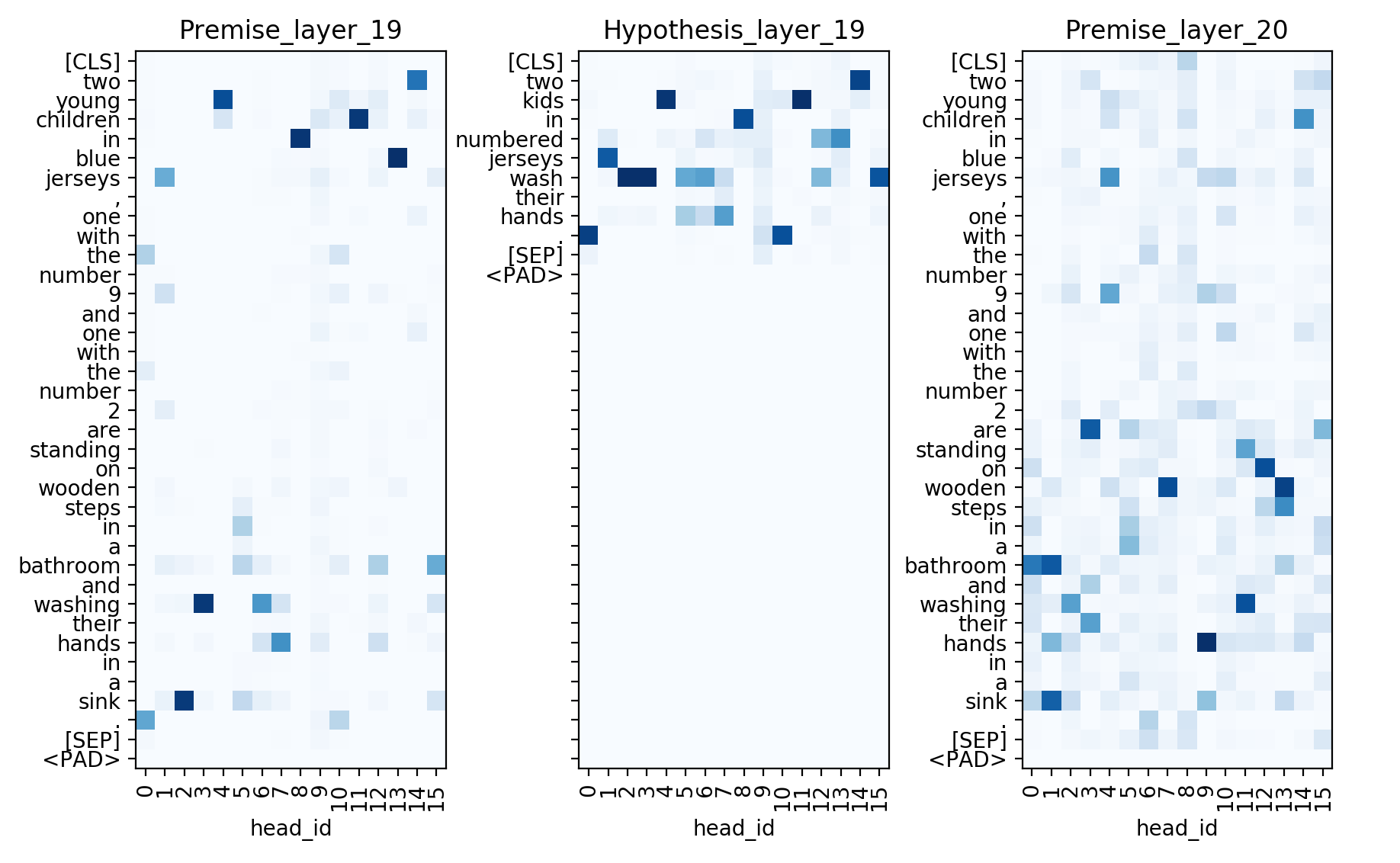}
\caption{Visualization on the attention weights of different heads (16 in total) from layer 19 and 20 on two sentences. }
\label{fig:visatt}
\end{figure*}

\paragraph{Effect of Number of Layers}
Next, we analyze the effects of the number of Transformer layers on our models, as shown in the Figure~\ref{fig:headlayers}.
We compare three different settings, i.e.,  ``Classic Match", ``Single-level Head-wise Match", and ``Multi-level Head-wise Match".
We observe that the head-wise matching based methods can always achieve better performance than the classic match method, regardless of how many Transformer layers we use.
We also observe that the more number of Transformer layers may not always get better performance for all the match models.
Lastly, our ``Multi-level Head-wise Match" model dynamically integrates all the useful head-wise matching representations from different layers.
As a result, this model achieves better performance than only using the representations from the final layer.

\paragraph{Visualization} 
Finally, to further explain our motivation on head-wise match, we conduct a visualization study on the attention weights of Transformer, as shown in Figure~\ref{fig:visatt}.
We observe the attention weights are not uniformly distributed. 
For example, the second head in the ``Premise\_layer\_19" focuses on the word ``jerseys" and the fifth head pays attention on the word ``young".
The head with the same id from different encoders can focus on the related words.
For example, the second head from ``Premise\_layer\_19" and ``Hypothesis\_layer\_19" focus on the ``jerseys", and the forth head focus on the ``young" and ``kids" respectively.
As a result, head-wise matching can help identify which aspects of the hypothesis can be explained by the premise. Moreover, the aggregation of these aspect matching will lead to the final sequence matching.

Additionally, we also observe that some heads pay attention on the punctuations, such as the first head in ``Premise\_layer\_19" and ``Hypothesis\_layer\_19". 
The maxpooling aggregation function learns to filter these types of head matching that will not contribute to the final prediction.
Finally, we can also see that the heads from different layers focus on different aspects of the sequence.
For example, the distribution of the words drawn attention by the heads from ``Premise\_layer\_19" and ``Premise\_layer\_20" are quite different. 
Hence, this explains the need for the head-matching across different layers.

\section{Conclusions}
In this paper, we focused on sequence matching based on pre-computed vector representations. 
We provide a comprehensive deep analysis on the representations built by pretrained Transformer models, making a key observation that merging all the heads to build sequence representation for matching is sub-optimal.
Instead, we propose to integrate the head-wise match between sequences, achieving a substantial performance gain on 5 different sequence matching tasks.
Moreover, by integrating the head-wise match from all different Transformer layers, our model achieves the best performance among the sentence vector-based models.
\bibliography{aaai2019}
\bibliographystyle{aaai}

\end{document}